%% file: main.tex
\documentclass[10pt,twocolumn,letterpaper]{article}

\usepackage{cvpr}
\usepackage{times}
\usepackage{epsfig}
\usepackage{graphicx}
\usepackage{amsmath}
\usepackage{amssymb}
\usepackage{caption}
\usepackage{subcaption}
\usepackage{multirow}
\usepackage{verbatim}

\usepackage[pagebackref=true,breaklinks=true,letterpaper=true,colorlinks,bookmarks=false]{hyperref}

 \cvprfinalcopy 

\def\cvprPaperID{2976} 

\ifcvprfinal\fi
\begin{document}

\title{Deep Residual Network for Joint Demosaicing and Super-Resolution}

\author{Ruofan Zhou, Radhakrishna Achanta, Sabine S{\"u}sstrunk\\
IC, EPFL\\
{\tt\small {\{ruofan.zhou, radhakrishna.achanta, sabine.susstrunk}\}@epfl.ch}
}

\maketitle

\input{texinput/abstract}
\input{texinput/introduction}
\input{texinput/relatedwork}

\input{texinput/ourmethod}
\input{texinput/experiments}
\input{texinput/conclusion}

{\small
\bibliographystyle{ieee}
\bibliography{egbib}
}

\end{document}

%% file: texinput/abstract.tex
\begin{abstract}
In digital photography, two image restoration tasks have been studied extensively and resolved independently: demosaicing and super-resolution. Both these tasks are related to resolution limitations of the camera. Performing super-resolution on a demosaiced images simply exacerbates the artifacts introduced by demosaicing. In this paper, we show that such accumulation of errors can be easily averted by jointly performing demosaicing and super-resolution. To this end, we propose a deep residual network for learning an end-to-end mapping between Bayer images and high-resolution images. By training on high-quality samples, our deep residual demosaicing and super-resolution network is able to recover high-quality super-resolved images from low-resolution Bayer mosaics in a single step without producing the artifacts common to such processing when the two operations are done separately. We perform extensive experiments to show that our deep residual network achieves demosaiced and super-resolved images that are superior to the state-of-the-art both qualitatively and in terms of PSNR and SSIM metrics.
\end{abstract}

%% file: texinput/introduction.tex
\section{Introduction}
\input{texinput/figureIntro1}

There is an evergrowing interest in capturing high-resolution images that is in step with the increasing quality of camera sensors and display devices. Ironically, the most prevalent image capture devices are mobile phones, which are equipped with small lenses and compact sensors. Despite the large advancements made in improving the dynamic range and resolution of images captured by mobile devices, the inherent design choices limit the ability to capture very high-quality images over the last decade.

The limitations come from two design issues. Firstly, the single CMOS sensor in most of the cameras, including mobile cameras, measures at each spatial location only a limited wavelengths range (red, green \emph{or} blue) of the electromagnetic radiation instead of the full visible spectrum (red, green, \emph{and} blue). This is achieved by placing a color filter array (CFA) in front of the sensor. The most common type of CFA is the Bayer pattern, which captures an image mosaic with twice green for each red and blue waveband. Secondly, as the sensor needs to be compact to fit into the device, resolution is limited by the size of the photon wells. Small photon wells have a low well capacity, which limits the dynamic range of the image capture. Large photon wells limit the number of pixels and thus resolution. To reconstruct full color from the CFA mosaiced image, demosaicing algorithms are applied, while low resolution can only be dealt with using super-resolution algorithms in a post-processing step.

In the last few decades, demosaicing and super-resolution have been independently studied and applied in sequential steps. However, the separate application of demosaicing and super-resolution is sub-optimal and usually leads to error accumulation. This is because artifacts such as color zippering introduced by demosaicing algorithms is treated as a valid signal of the input image by the super-resolution algorithms. As most of the super-resolution algorithms~\cite{Dong:2014vh} rely on the assumption that the human visual system is more sensitive to the details in the luminance channel than the details in chroma channels, they only deal with noise in the luminance channel, which neglects the artifacts in chroma channels caused by demosaicing algorithms. As a result, sequential application of super-resolution algorithms after demosaicing algorithms leads to visually disturbing artifacts in the final output. (see examples in Figure.~\ref{fig:intro}).

The algorithms for demosaicing and super-resolution are meant to overcome the sampling limitations of digital cameras. While they have been dealt with in isolation, it is reasonable to address them in a unified context, which is the aim of this paper. With the advent of deep learning, there are several methods for super-resolution~\cite{Dong:2014vh, Dong:2016ek, Kim:2016kv, Kim:2016wv, Lim:2017un,   Shi:2016tm} that successfully outperform traditional super-resolution methods~\cite{Elad:2010wu, Egiazarian:2015ww,  Smith:2012hv, Freedman:2011cf, Sun:2008ia, Timofte:2015kw, Yang:2012db}. Only recently, deep learning has also used successfully for image demosaicing~\cite{Durand:2016tz}. When using deep learning, it is possible to address demosaicing and super-resolution simultaneously, as we show in this paper.

\subsection{Contributions}
Unlike previous works, in this paper we propose to use a deep residual network for end-to-end joint demosaicing and super-resolution. More specifically, our network can learn an end-to-end mapping between RGGB Bayer patterns and high-resolution color images. This network generalizes to other color filter arrays (CFA) with a simple modification of 2 layers of the network.

To the best of our knowledge, ours is the first attempt to perform joint demosaicing and super-resolution. Unlike existing super-resolution methods that usually super-resolve only the luminance channel (while resorting to interpolation of the chroma channels), we generate full-color three channel super-resolution output directly.

Since both demosaicing and super-resolution are jointly optimized through the network, conventional artifacts such as moir\'e and zippering, which pose a post-processing challenge, are nearly eliminated.

We demonstrate both quantitatively and qualitatively that our approach generates higher quality results than state-of-the-art. In addition, our method is computationally more efficient because of the joint operation. Our approach can be extended to videos, and can potentially be integrated into the imaging pipeline.

%% file: texinput/figureIntro1.tex
\begin{figure}
    \begin{tabular}{cc}
        \multicolumn{2}{c}{\includegraphics[trim={5mm 0 0 0},clip, width=.95\linewidth]{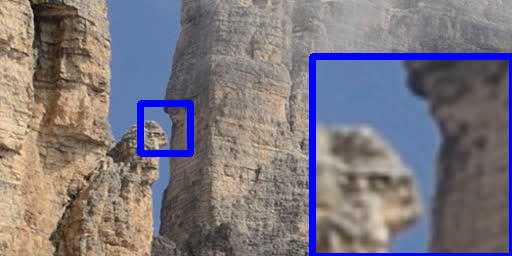}}\\
        \multicolumn{2}{c}{Reference image from RAISE~\cite{DangNguyen:2015va}}\\
        \includegraphics[width=.45\linewidth]{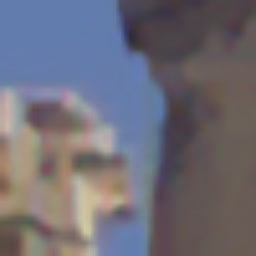} & \includegraphics[width=.45\linewidth]{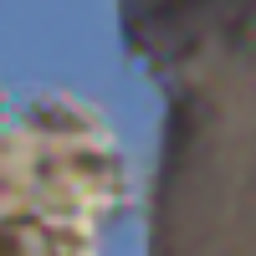} \\
        \footnotesize{ADMM\cite{Klatzer:2016wg}+SRCNN\cite{Dong:2014vh}}&\footnotesize{FlexISP\cite{Pulli:2014gq}+SRCNN\cite{Dong:2014vh}}\\
        \footnotesize{(27.9406 dB,0.8082)}&\footnotesize{(28.2927 dB,0.8420)}\\
        \includegraphics[width=.45\linewidth]{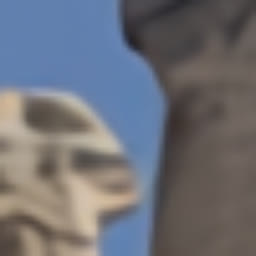} &
        \includegraphics[width=.45\linewidth]{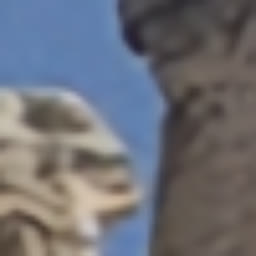}\\
        \footnotesize{DemosaicNet\cite{Durand:2016tz}+SRCNN\cite{Dong:2014vh}} & \footnotesize{Our output}\\
        \footnotesize{(30.1896 dB,0.8920)} & \footnotesize{{(30.9535 dB,0.9118)}}
    \end{tabular}
    \caption{Comparison of our joint demosaicing and super-resolution output to the state-of-the-art. The two numbers in the brackets are PSNR and SSIM, respectively. Note how the sequential application of demoisacing and super-resolution carries forward color artifacts (second row) or blurring (third row, left). Our method exhibits none of these artifacts and is able to faithfully reconstruct the original.}
        \label{fig:intro}
\end{figure}

%% file: texinput/relatedwork.tex
\section{Related work}
Our goal of joint demosaicing and super-resolution is to directly recover a high-resolution image from a low-resolution Bayer pattern. Both demosaicing and super-resolution are well-studied problems. Since we are the first to address the joint solution, in this section we briefly present the traditional literature that deal with these two problems independently. 


\begin{figure*}
    \centering
    \includegraphics[width=.95\linewidth]{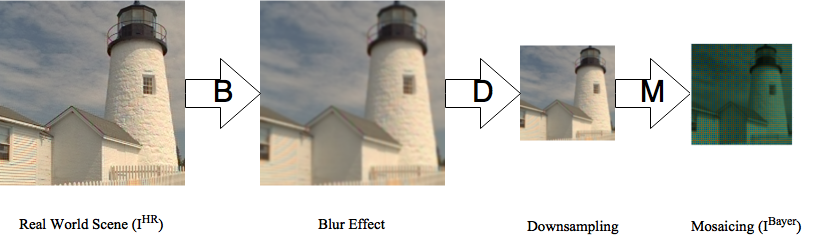}
    \caption{Block diagram presenting the assumed image formation in our model. Where $I^{HR}$ is the intensity distribution of the real scene, $B$, $D$, $M$ present the blurring, downsampling and mosaicing process, $I^{Bayer}$ is the observed Bayer image.}
    \label{fig:formation}
\end{figure*}

\subsection{Demosaicing}
To reduce manufacturing costs, most camera sensors capture only one of red, green, or blue channels at each pixel~\cite{Farsiu:2006ug}. This is achieved by placing a color-filter array (CFA) in front of the CMOS. The Bayer pattern is a very common example of such a CFA. Demosaicing is the process to recover the full-color image from the incomplete color samples output from this kind of image sensor, which is the crucial first step of most digital camera pipelines.

To solve this problem, early approaches use different interpolations for luminance channel and chrominance channel in spatial domain~\cite{Wu:hw, Li:2008ic} or frequency domain~\cite{Alley:2005ez, Glotzbach:jm} to fill the missing pixels. However, these methods introduce artifacts like zippering and false color artifacts. Some methods resort to post-processing approaches such as median filtering~\cite{Hirakawa:cy} to mitigate these artifacts at the cost of introducing other artifacts.

Advanced works tend to build the demosaicing on the underlying image statistics. These methods rely on techniques ranging from SVM regression to shallow neural network architectures. These methods outperform the traditional methods and give the state-of-the-art result. Heide \etal~\cite{Pulli:2014gq} formulate the demosaicing as an image reconstruction problem and embed a non-local natural image prior to an optimization approach called FlexISP to achieve natural results. Klatzer \etal~\cite{Klatzer:2016wg} build a variational energy minimization framework SEM to efficiently learn suitable regularization term from training data, thus yield high-quality results in the presence of noise. More recently, Gharbi \etal~\cite{Durand:2016tz} proposed a deep learning-based demosaicing method (DemosaicNet) to improve the quality of demosaicing by training on a large dataset.

\subsection{Single Image Super-Resolution}
Single image super-resolution aims to reconstruct a high-resolution image from a single low-resolution image. Traditional interpolation approaches based on sampling theory~\cite{Allebach:1996uz, :2001uc} has encountered limitations in producing realistic details. Recent methods~\cite{Egiazarian:2015ww, Elad:2010wu} tend to constrain this ill-posed problem by embedding prior knowledge from large datasets. While these data-driven methods were using simple architectures and hard-coded heuristics, they do not compare favorably with the recent state-of-the-art, which relies on CNN's for super-resolution.

Inspired by the success of CNN in image classification tasks~\cite{Krizhevsky:2012wl}, various CNN architectures have been proposed for single image super-resolution~\cite{Dong:2014vh, Dong:2016ek, Kim:2016kv, Kim:2016wv, Mao:2016ti}. By using modern neural network techniques such as skip-connections~\cite{mao:2016} and residual blocks~\cite{He:2016tt}, these networks alleviate the burden of carrying identity information in the super-resolution network. They have significantly improved the performance of super-resolution in terms of peak signal-to-noise ratio (PSNR).

Note that almost all modern super-resolution methods have been designed to increase the resolution of a single channel (monochromatic) image, they usually only focus on the luminance channel in the YCbCr color space as human eyes are more sensitive to luminance changes~\cite{Schulter:2015tz}. These methods are sub-optimal as they do not fully exploit the correlation across the color bands. These methods may generate poor quality outputs when color artifacts are inherited from the lower-resolution images (Figure. ~\ref{fig:intro}).

In our work, the joint addressing of demosaicing and super-resolution results in fewer visual artifacts, higher PSNR, and at the same time, lower computational cost.

%% file: texinput/ourmethod.tex
\section{Joint Demosaicing and Super-Resolution}

A common image formation model for imaging systems is illustrated in Figure.~\ref{fig:formation}. In this model, the real world scene $I^{HR}$ is smoothed by a blur kernel which respects to the point spread function of the camera, then it is downsampled by a factor of $r$ and mosaiced by the CFA by the CMOS to get the observed Bayer image $I_{Bayer}$. Our goal is to provide an approximate inverse operation estimating a high-resolution image $I^{SR} \approx I^{HR}$ given such a low-resolution Bayer image $I^{Bayer}$. In general, $I^{Bayer}$ is a real-valued tensor of size $H \times W \times 1$, while $I^{HR}$ is a tensor of $r\cdot H \times r\cdot W \times 3$. This problem is highly ill-posed as the downsampling and mosaicing are non-invertible.

To solve this, traditional methods usually design nonlinear filters that incorporate prior heuristics about inter- and intra-channel correlation. A deep CNN is a better substitute for such methods, as convolutional layers can automatically learn to exploit inter- and intra-channel correlation through a large dataset of training images. Moreover, the exclusive use of a set of convolutional layers enables joint optimization of all the parameters to minimize a single objective as is the case in joint demosaicing and super-resolution.

We thus build our framework in a data-driven fashion: we create the training set from a large set of high-quality images $I^{HR}$, and produce the input measurements $I^{Bayer}$ using the same process as the image formation model illustrated in Figure.~\ref{fig:formation}, then we train our deep convolutional network on this dataset.  

\begin{figure*}[]
    \centering
    \includegraphics[width=.95\linewidth]{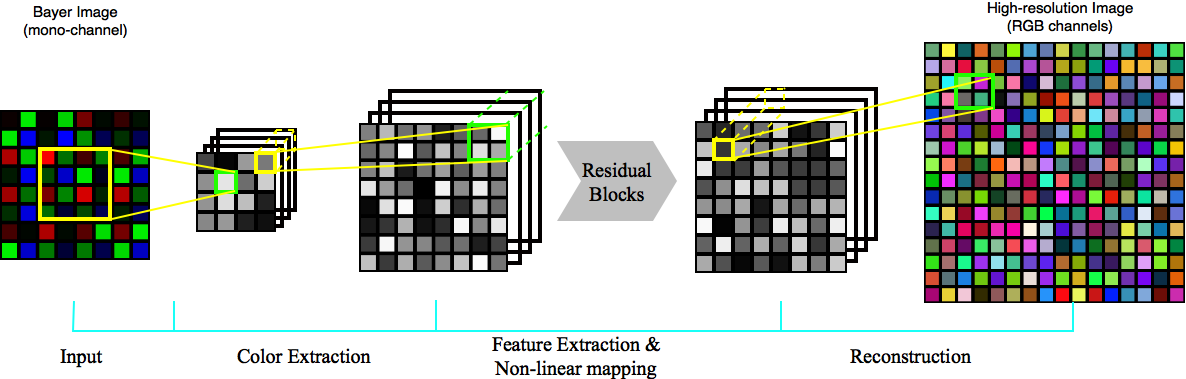}
    \caption{Illustration of our proposed network architecture. The network is a feed-forward fully-convolutional network that maps a low-resolution Bayer image to a high-resolution color image. Conceptually the network has three components: color extraction of Bayer image, non-linear mapping from Bayer image representation to color image representation with feature extraction, and high-resolution color image reconstruction.}
    \label{fig:network}
\end{figure*}

\subsection{Deep Residual Network Design}

\begin{table}[]
    \centering
    \begin{tabular}{|c|c|c|}
        \hline
        Stage & Layer & Output Shape  \\
        \hline
        \multicolumn{2}{|c|}{Input (Bayer image)} &  $h \times w \times 1$ \\
        \hline
        1 & Conv with a stride of 2 & $\frac{h}{2} \times \frac{w}{2} \times C$ \\
          & Sub-pixel Conv & $h \times w \times \frac{C}{4}$ \\
          & Conv, PReLU & $h \times w \times C$\\
        \hline
        2 & Residual Block & $h \times w \times C$ \\
          & ... & $h \times w \times C$  \\
          & Residual Block & $h \times w \times C$  \\
        \hline
        3 & Sub-pixel Conv & $2\cdot h \times 2\cdot w \times \frac{C}{4}$ \\
          & Conv, PReLU & $2\cdot h \times 2\cdot w \times C$ \\
          & Conv & $2\cdot h \times 2\cdot w \times 3$ \\
        \hline
        \multicolumn{2}{|c|}{Output (color image)} & $2\cdot h \times 2\cdot w \times 3$\\
        \hline
    \end{tabular}
    \caption{The summary of our network architecture. The stages $1, 2, 3$ of the first column correspond to the three stages of color extraction, feature extraction \& non-linear mapping, and reconstruction, respectively) illustrated in Figure.~\ref{fig:network}. We set the number of filters $C = 256$ and use 24 residual blocks in stage 2.}
    \label{tab:network}
\end{table}
We use a standard feed-forward network architecture to implement our joint demosaicing and super-resolution, which is presented in Figure~\ref{fig:network}. The goal of the network is to recover from $I^{Bayer}$ an image $I^{SR} = F(I^{Bayer})$ that is as similar as possible to the ground truth high-resolution color image $I^{HR}$. We wish to learn a mapping $F$ from a large corpus of images, which conceptually consists of three stages:
\begin{enumerate}
  \item \textbf{Color Extraction:} this operation separates the color pixels into different channels from the mono-channel Bayer image. With this operation, no hand-crafted rearrangement of the Bayer input is needed unlike other demosaicing algorithms~\cite{Tan:2017, Pulli:2014gq, Durand:2016tz}. This operation gives a set of color features from the Bayer input.
  \item \textbf{Feature Extraction \& Non-linear Mapping:} following the intuition of building the first deep neural network for super-resolution~\cite{Dong:2014vh}, this operation extracts overlapping patches from the color features to use high-dimensional vectors to represent the Bayer image in a low-resolution manifold,  which is then mapped to the high-resolution manifold. 
  \item \textbf{Reconstruction:} this operation aggregates high-resolution representations to generate the final high-resolution color image $I^{SR}$.
\end{enumerate}

\subsubsection{Color Extraction}

The Bayer image is a matrix with the three color channel samples arranged in a regular pattern in a single channel. To make the spatial pattern translation-invariant and reduce the computational cost in latter steps, it is essential to separate the colors in the Bayer image into different channels at the beginning. The Bayer pattern is regular and has a spatial size of $s \times s$, where $s = 2$ and since the neighboring colors may also affect the result, we build our first convolutional layer $L_{1}$ with a spatial size of $2 \cdot s$ and a stride of $s$:
\begin{equation}
I^{1} = L_{1}(I^{Bayer})_{(x,y)} = (W_1 * I^{Bayer} + b_1)_{(2\cdot x, 2\cdot y)},
\end{equation}
where $I^1$ represents the output from the first layer, $W_1$ and $b_1$ represent the filters and biases of the first convolutional layer, and $*$ denotes the convolution operation. Here, $W_1$ corresponds to $C = 256$ filters of support $2 \cdot s \times 2 \cdot s$.

We build an efficient sub-pixel convolutional Layer~\cite{Shi:2016tm} $L_{2}$ to upsample the color features back to the original resolution:
\begin{equation}
L_{2}(I^{1})_{(x,y,c)}=I^{1}_{(\left \lfloor \frac{x}{s} \right \rfloor, \left \lfloor \frac{y}{s} \right \rfloor, \frac{C \dot mod(y,s)}{s}+\frac{C \cdot mod(x,s)}{s^2}+c)},
\end{equation}
\noindent here, the sub-pixel convolutional layer is equivalent to a shuffling operation which reshapes a tensor of size $H \times W \times C$ into a tensor of size $s \cdot H \times s\cdot W \times \frac{C}{s^2}$. We find that applying this sub-pixel convolutional layer helps reduce checkerboard artifacts in the output.

Due to the linearity of the separation operation, no activation function is utilized in either layer. Note that the color extraction operation can be generalized to other CFAs by modifying $s$ respect to the spatial size and arrangement of the specific CFA. Thus we have $s = 2$ for all kinds of Bayer CFAs, CYGM CFA or RGBE CFA, and $s = 6$ for the X-trans pattern~\cite{xtrans}.

\subsubsection{Feature Extraction \& Non-linear Mapping}
\begin{figure}
    \centering
    \includegraphics[width=.95\linewidth]{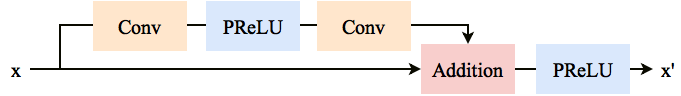}
    \caption{Illustration of the architecture of our residual blocks. We remove the batch normalization layer in the original residual blocks~\cite{He:2016tt} and replace the ReLU with Parametric ReLU~\cite{He:2015cv}. This structure enables faster convergence and better performance.}
    \label{fig:residual}
\end{figure}
Inspired by Dong \etal~\cite{Dong:2014vh}, to explore relationships within each color channel and between channels, as well as to represent the Bayer image in a high-resolution manifold, we exploit a group of convolutional layers in this step.

Previous works~\cite{He:2016tt} have demonstrated that residual networks exhibit excellent performance both in accuracy and training speed in computer vision problems ranging from low-level to high-level tasks. We build a set of $n_b$ residual blocks each having a similar architecture as Lim \etal~\cite{Lim:2017un}, which is demonstrated in Figure.~\ref{fig:residual}. We remove the batch normalization layers in the original residual blocks~\cite{He:2016tt} since these layers get rid of range flexibility from networks by normalizing the features~\cite{Lim:2017un}. We also replace the activation functions ReLU with Parametric ReLU~\cite{He:2015cv}(PReLU) for preventing dead neurons and vanishing gradients caused by ReLU. These modifications help stabilize the training and reduce color shift artifacts in the output. For convenience, we set all residual network blocks to have the same number of filters $C = 256$.

\subsubsection{Reconstruction}
In the reconstruction stage, we apply another sub-pixel convolutional layer to upsample the extracted features to the desired resolution. This is followed by a final convolutional layer to reconstruct the high-resolution color image.

%% file: texinput/experiments.tex
\section{Experiments}
\subsection{Datasets}
\begin{figure}
    \centering
    \includegraphics[width=.95\linewidth]{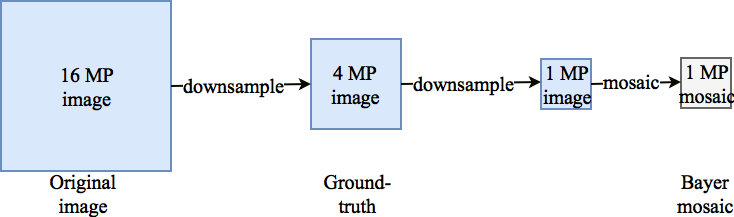}
    \caption{Illustration of the steps we take to create the input and output images of our training and testing dataset. The original 16 megapixel images are downsized to 4 megapixel eliminate demosaicing errors. The 4 megapixel images serve as reference super-resolution images, whose downsampled 1 megapixel version provide the the single-channel Bayer CFA images used as input to our network.}
    \label{fig:dataset}
\end{figure}

For training and evaluation of the network, we use publicly available dataset RAISE~\cite{DangNguyen:2015va} which provides 8,162 uncompressed raw images as well as their demosaiced counterparts in TIFF format.

It is to be noted that if we use images that are already demosaiced by a given algorithm to our network, the network will learn to generate any artifacts introduced by the demosaicing algorithm. We circumvent this problem as follows. We use the demosaiced images of RAISE that are larger than 16 megapixels in size. We then perform a progressive downsizing of the image in steps by a factor of 1.25 each time until we obtain one-fourth of the original image size (i.e down to about 4 megapixels). This is done to eliminate artifacts that have potentially been introduced by the demosaicing algorithm as well as by other factors in the camera processing pipeline (like sensor noise). This way we obtain high-quality ground-truth $I^{HR}$, to serve as the super-resolved images.

To create input Bayer images $I^{Bayer}$ from these ground-truth images, we further downsample the previously downsample images to one-fourth of the size (to about 1 megapixels). We follow the assumed image formation demonstrated in Figure. ~\ref{fig:formation}. As required for the Bayer pattern, we set the downsample factor $r = 2$, and sample pixels from the three channels in the Bayer CFA pattern to obtain a single-channel mosaiced images as low-resolution input images for training. Thus for a $H \times W \times 1$ Bayer image input, the desired color image output is of size $2 \cdot H \times 2\cdot W \times 3$.These steps are illustrated in Fig.~\ref{fig:dataset}.

To train our network, we use a subset of RAISE of 6,000 images. In particular, we randomly selected 4,000 photos from Landscape category and randomly selected 2,000 photos from other categories. We also randomly select 50 images from the rest of RAISE dataset to build the testing set.

\subsection{Training Details}
For training, we use $64 \times 64 \times 1$ sized patches from the created Bayer mosaics as input. As output images we use color image patches of size $128 \times 128 \times 3$ from the high-resolution (4 megapixel) images. We train our network with ADAM optimizer~\cite{Kingma:2014us} by setting learning rate $= 0.0001$, $\beta_{1} = 0.9$, $\beta_{2} = 0.999$, and $\epsilon = 10^{-8}$. We set mini-batch as 16. For better convergence of the network, we halve the learning rate after every $10000$ mini-batch updates.

\subsection{Results}
\begin{figure}[]
    \centering
    \includegraphics[width=.95\linewidth]{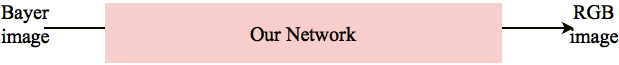}\\
    (a)\\
    \includegraphics[width=.95\linewidth]{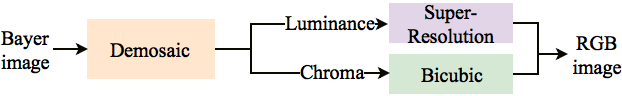}\\
    (b)
    \caption{(a) is our framework for joint demosaicing and super-resolution, our network can perform the whole process in an end-to-end manner. (b) shows a typical pipeline to combine the demosaic algorithms and super-resolution algorithms, which we use for comparing with other algorithms. Unlike most super-resolution algorithms that output only the luminance channel, we directly generate full color output.}
    \label{fig:diagram}
\end{figure}

\begin{figure*}
\centering
\begin{tabular}{cccc}
\multicolumn{4}{c}{Reference:}\\
\includegraphics[width=0.2\linewidth]{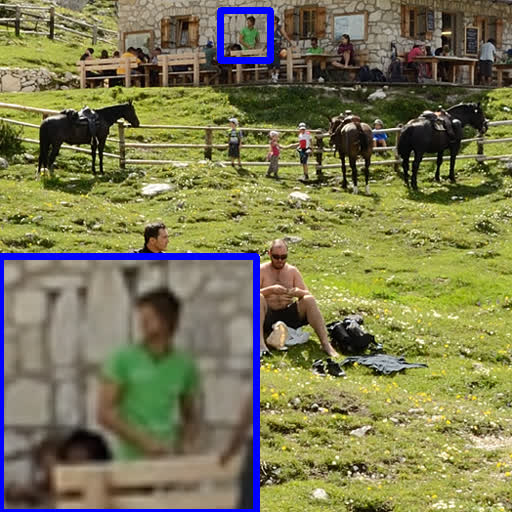}&
\includegraphics[width=0.2\linewidth]{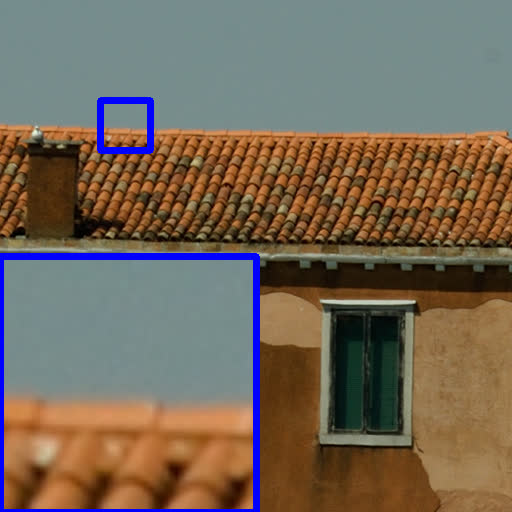}&
\includegraphics[width=0.2\linewidth]{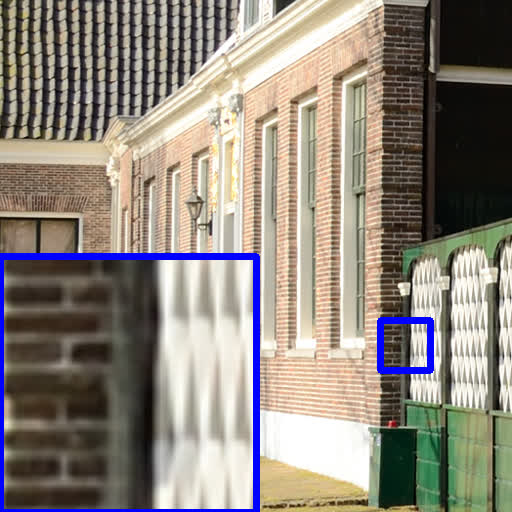}&
\includegraphics[width=0.2\linewidth]{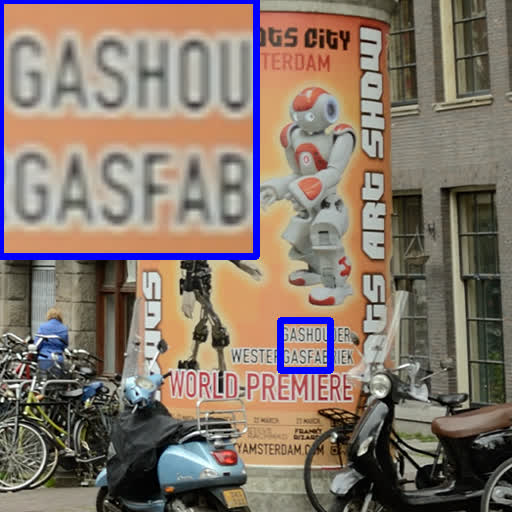}\\
\multicolumn{4}{c}{FlexISP~\cite{Pulli:2014gq} + SRCNN~\cite{Dong:2014vh}:}\\
\includegraphics[width=0.2\linewidth]{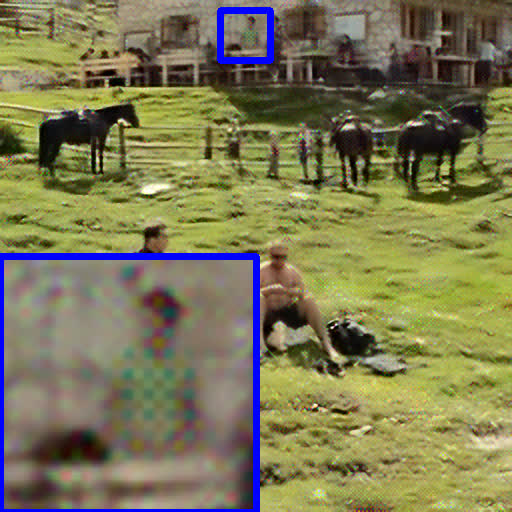}&
\includegraphics[width=0.2\linewidth]{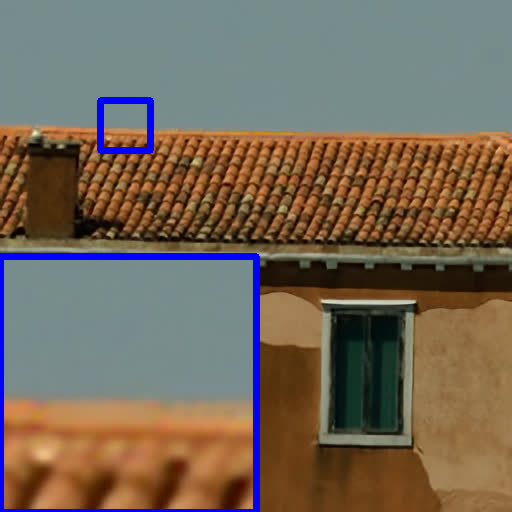}&
\includegraphics[width=0.2\linewidth]{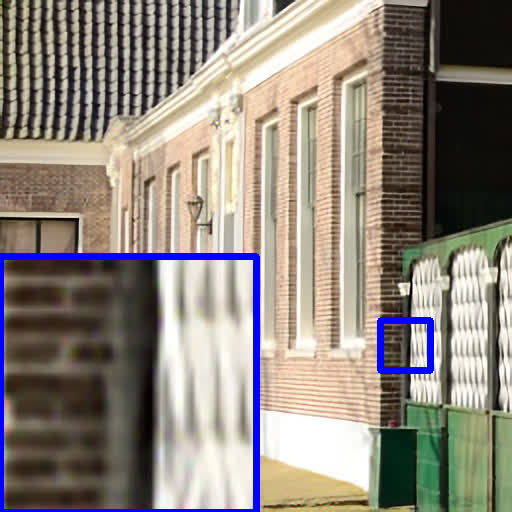}&
\includegraphics[width=0.2\linewidth]{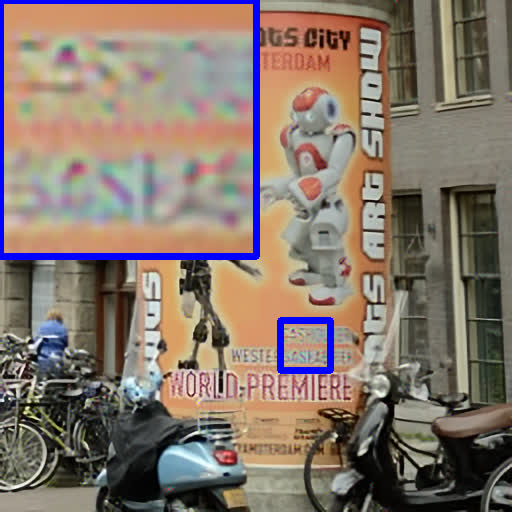}\\
(22.2315dB , 0.8855)&(32.9956dB , 0.9812)&(26.2782dB , 0.9133)&(27.5929dB , 0.9230)\\
\multicolumn{4}{c}{SEM~\cite{Klatzer:2016wg} + SRCNN~\cite{Dong:2014vh}:}\\
\includegraphics[width=0.2\linewidth]{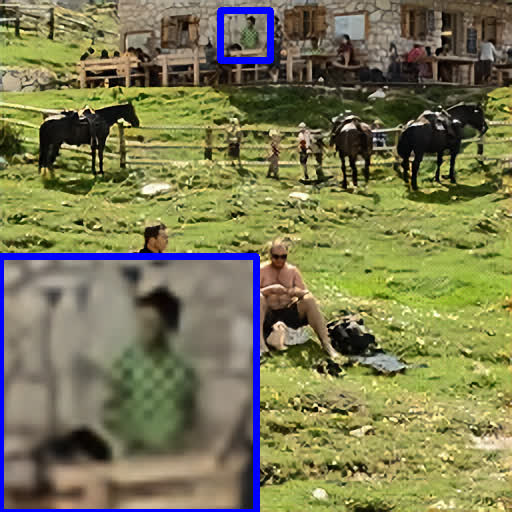}&
\includegraphics[width=0.2\linewidth]{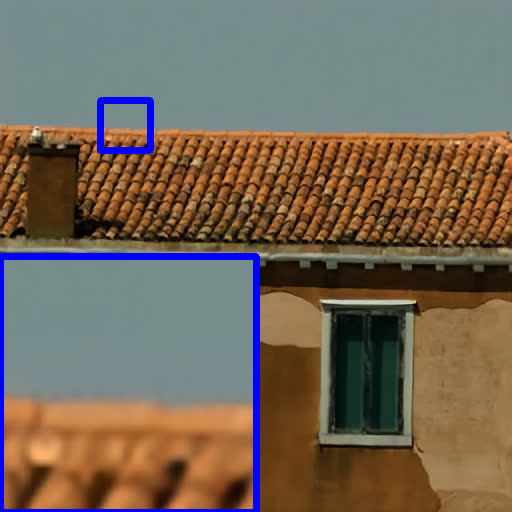}&
\includegraphics[width=0.2\linewidth]{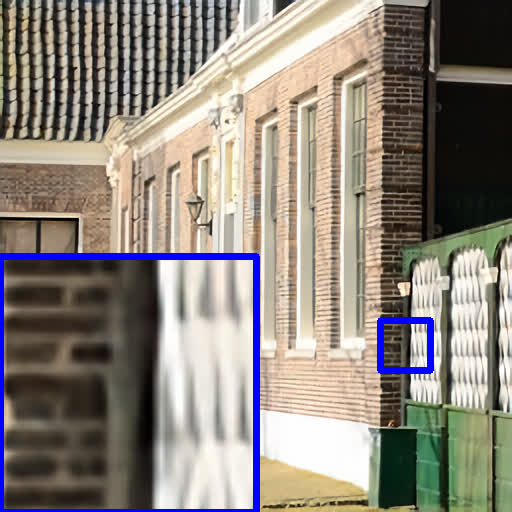}&
\includegraphics[width=0.2\linewidth]{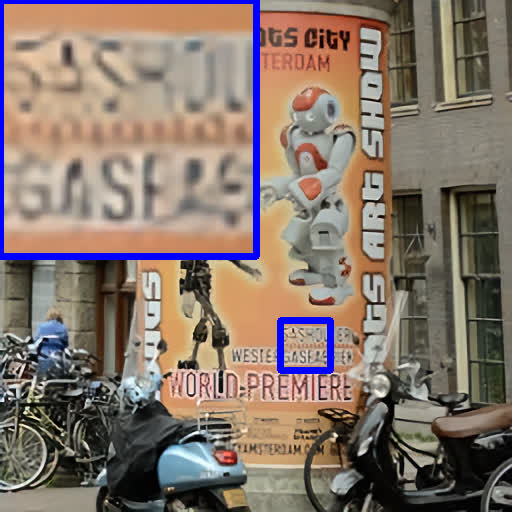}\\
(19.9758dB , 0.7898)&(32.5800dB , 0.9792)&(26.1953dB , 0.9153)&(27.5932dB , 0.9335)\\
\multicolumn{4}{c}{DemosaicNet~\cite{Durand:2016tz} + SRCNN~\cite{Dong:2014vh}:}\\
\includegraphics[width=0.2\linewidth]{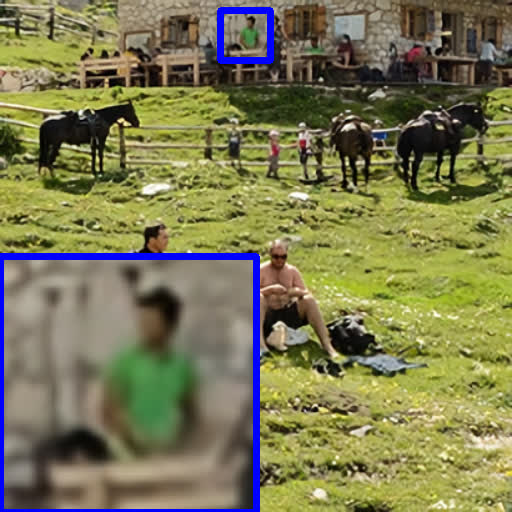}&
\includegraphics[width=0.2\linewidth]{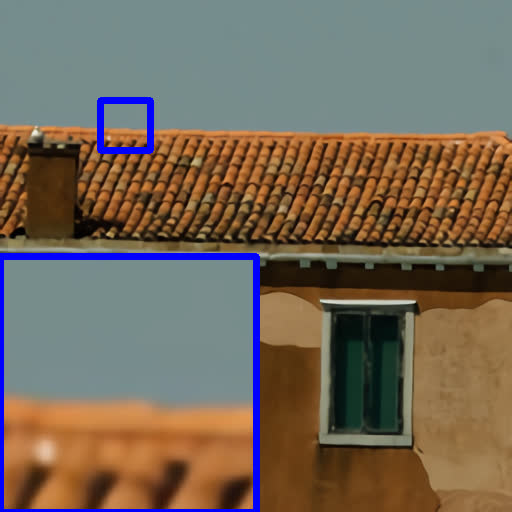}&
\includegraphics[width=0.2\linewidth]{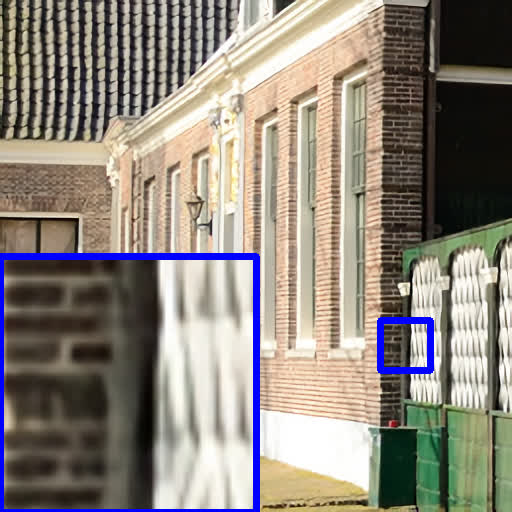}&
\includegraphics[width=0.2\linewidth]{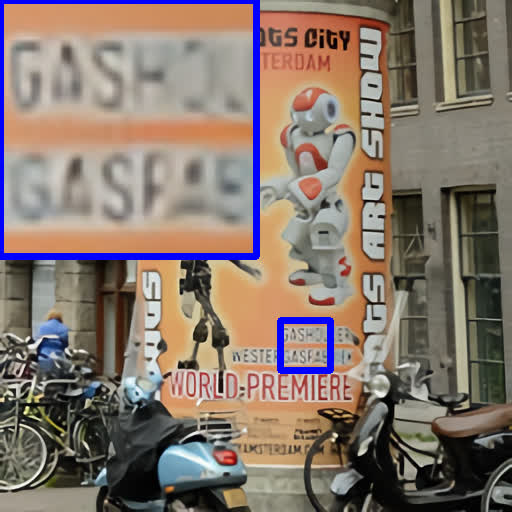}\\
(23.2090dB , 0.8995)&(32.3014dB , 0.9798)&(26.7680dB , 0.9261)&(28.5400dB , 0.9329)\\
\multicolumn{4}{c}{Ours:}\\
\includegraphics[width=0.2\linewidth]{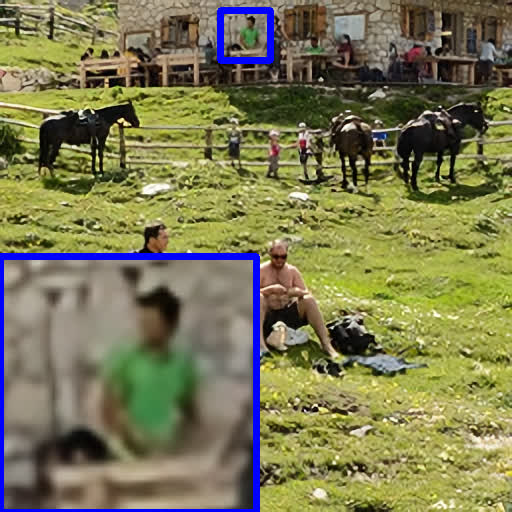}&
\includegraphics[width=0.2\linewidth]{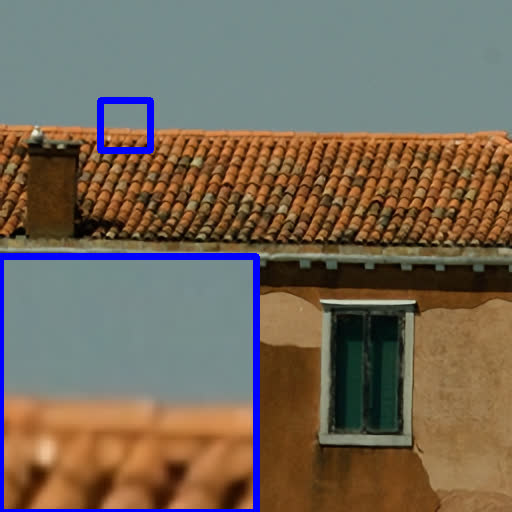}&
\includegraphics[width=0.2\linewidth]{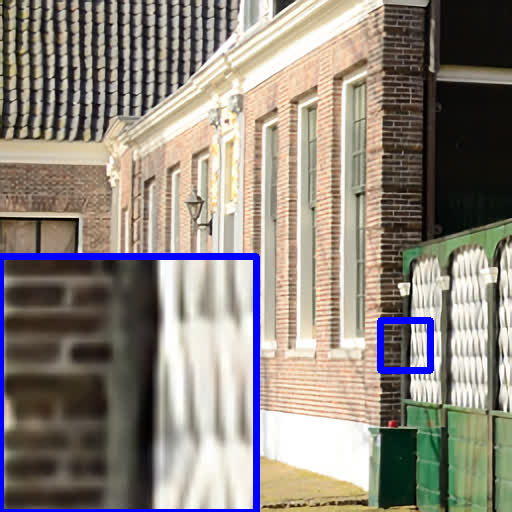}&
\includegraphics[width=0.2\linewidth]{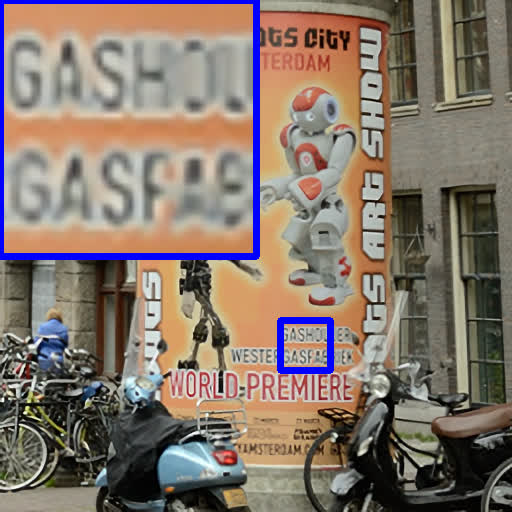}\\
(23.1523dB , 0.9052)&(34.2992dB , 0.9857)&(27.5472dB , 0.9382)&(29.6846dB , 0.9557)
\end{tabular}
\caption{Joint demosaicing and super-resolution results on images from the RAISE~\cite{DangNguyen:2015va} dataset. The two numbers in the brackets are the PSNR and SSIM scores, respectively.}
\label{fig:result1}
\end{figure*}

\begin{figure*}
\centering
\begin{tabular}{cccc}
\multicolumn{4}{c}{Reference:}\\
\includegraphics[width=0.2\linewidth]{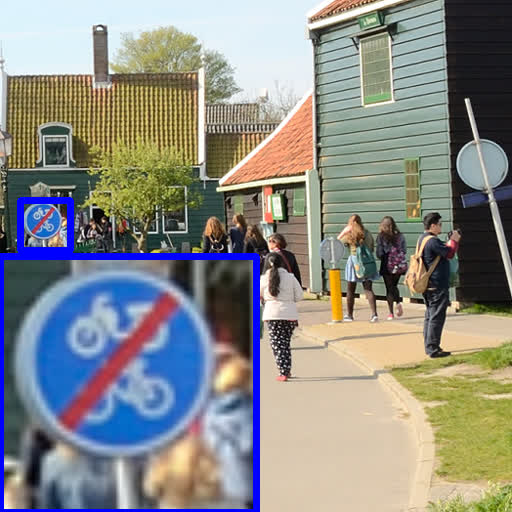}&
\includegraphics[width=0.2\linewidth]{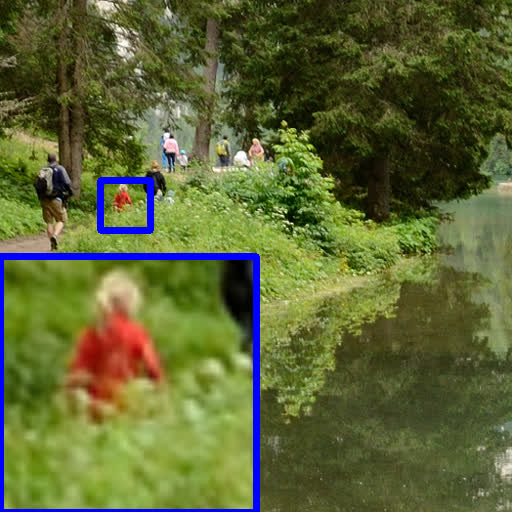}&
\includegraphics[width=0.2\linewidth]{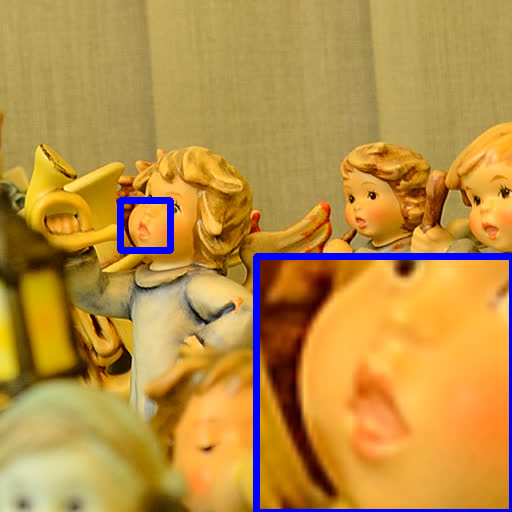}&
\includegraphics[width=0.2\linewidth]{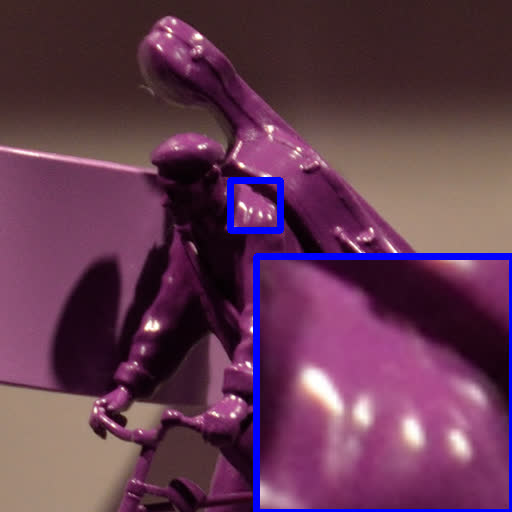}\\
\multicolumn{4}{c}{FlexISP~\cite{Pulli:2014gq} + SRCNN~\cite{Dong:2014vh}:}\\
\includegraphics[width=0.2\linewidth]{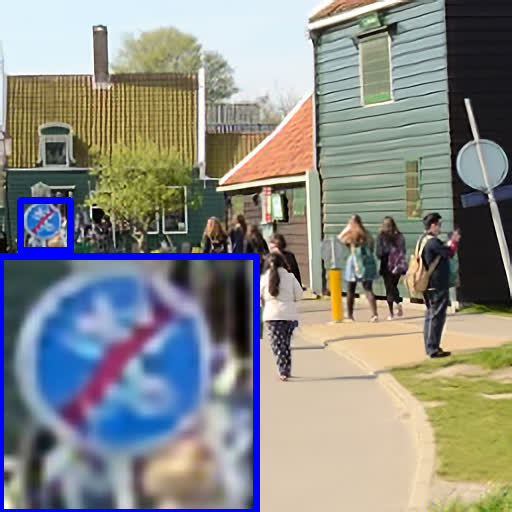}&
\includegraphics[width=0.2\linewidth]{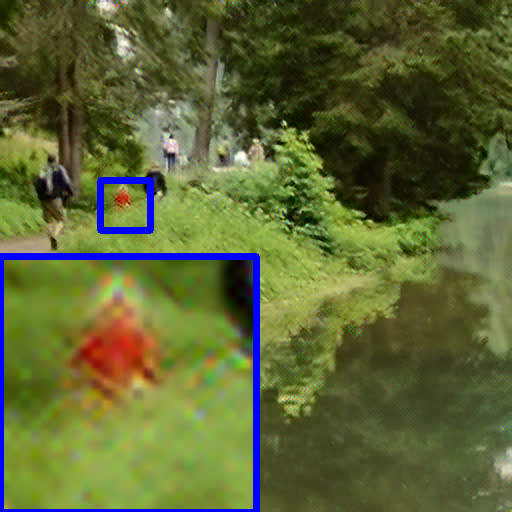}&
\includegraphics[width=0.2\linewidth]{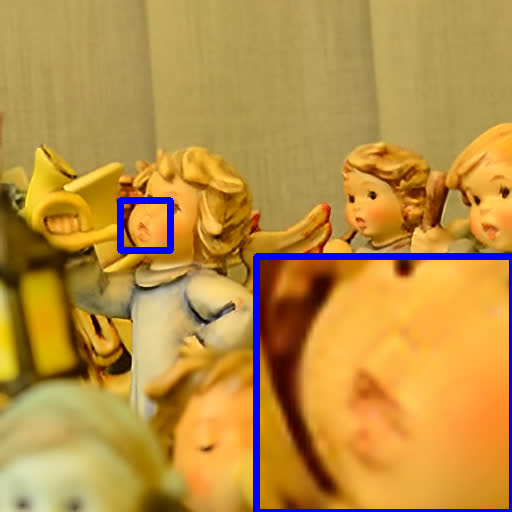}&
\includegraphics[width=0.2\linewidth]{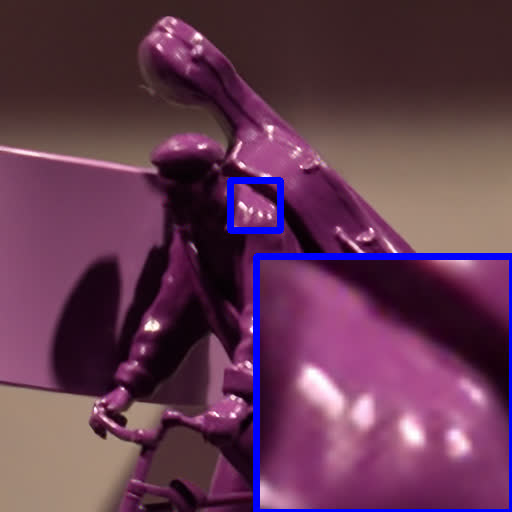}\\
(25.3831 dB , 0.9130)&(24.7385dB , 0.8852)&(32.3718dB , 0.9929)&(36.2369dB , 0.9805)\\
\multicolumn{4}{c}{SEM~\cite{Klatzer:2016wg} + SRCNN~\cite{Dong:2014vh}:}\\
\includegraphics[width=0.2\linewidth]{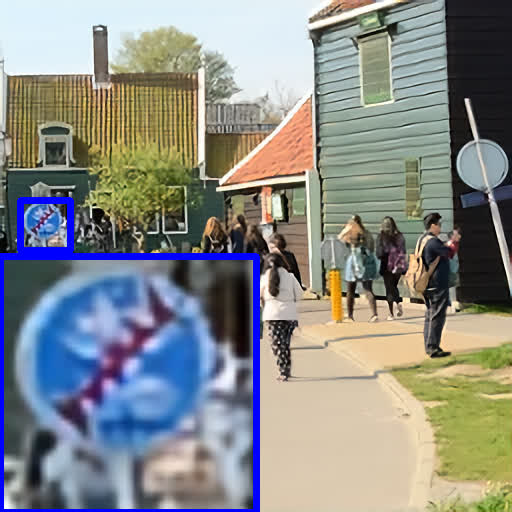}&
\includegraphics[width=0.2\linewidth]{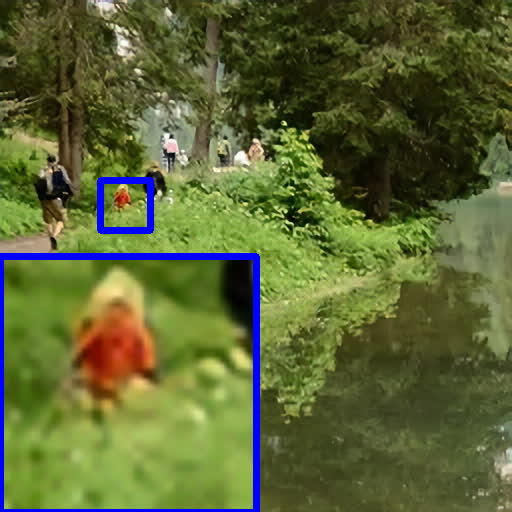}&
\includegraphics[width=0.2\linewidth]{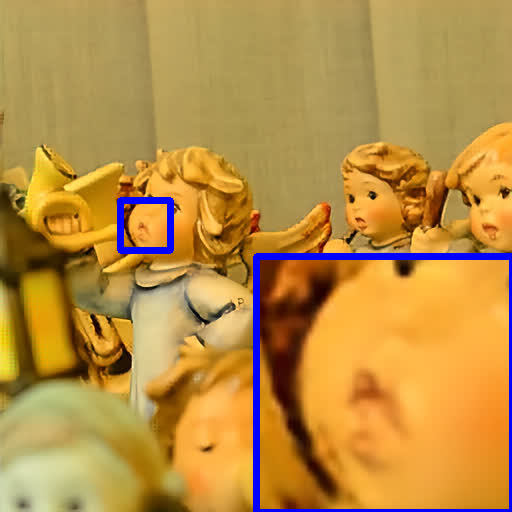}&
\includegraphics[width=0.2\linewidth]{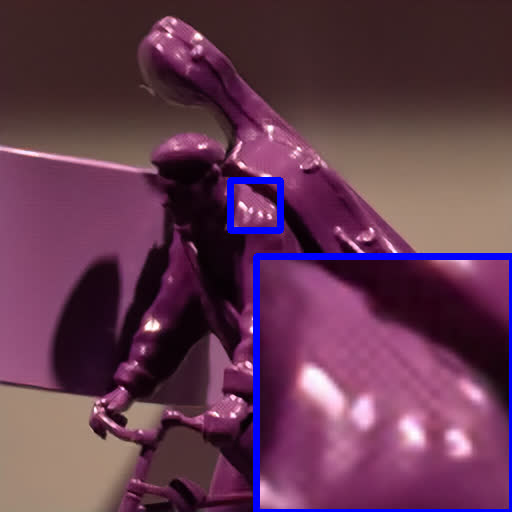}\\
(25.0950dB , 0.9131)&(27.1955dB , 0.9354)&(30.8814dB , 0.9900)&(34.6032dB , 0.9807)\\
\multicolumn{4}{c}{DemosaicNet~\cite{Durand:2016tz} + SRCNN~\cite{Dong:2014vh}:}\\
\includegraphics[width=0.2\linewidth]{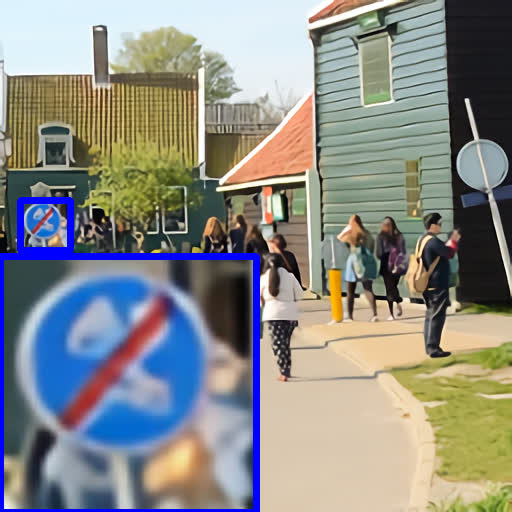}&
\includegraphics[width=0.2\linewidth]{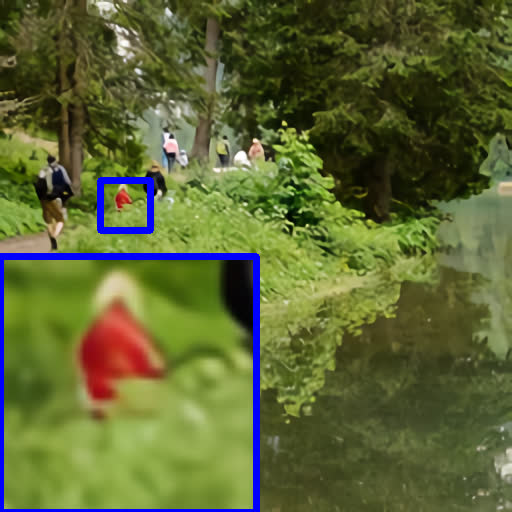}&
\includegraphics[width=0.2\linewidth]{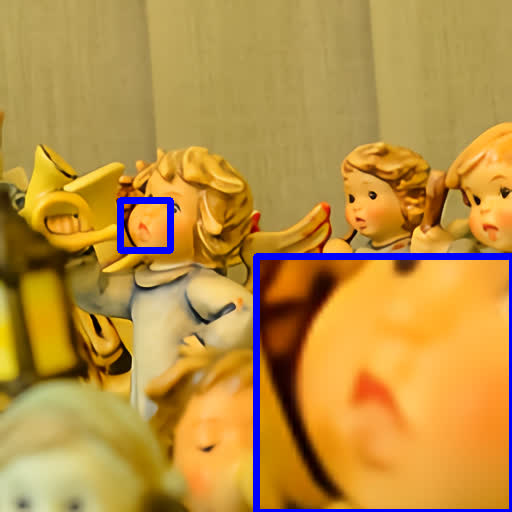}&
\includegraphics[width=0.2\linewidth]{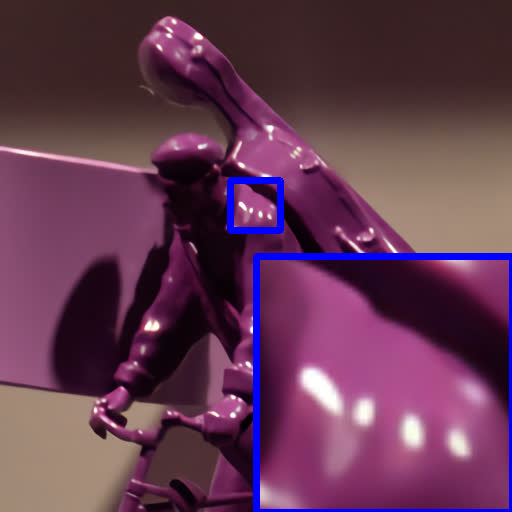}\\
(25.7484dB , 0.9206)&(27.4267dB , 0.9326)&(32.3960 dB , 0.9928)&(36.0122dB , 0.9829)\\
\multicolumn{4}{c}{Ours:}\\
\includegraphics[width=0.2\linewidth]{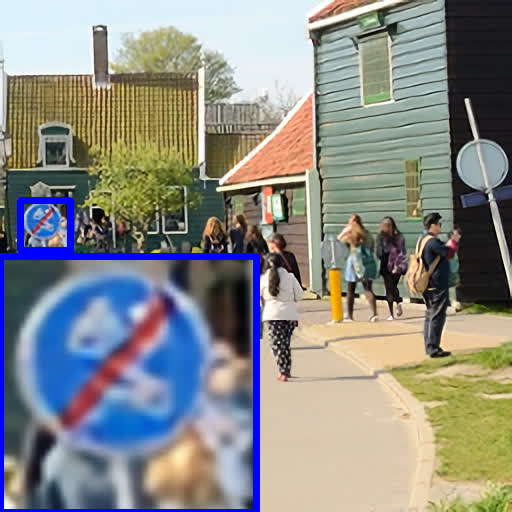}&
\includegraphics[width=0.2\linewidth]{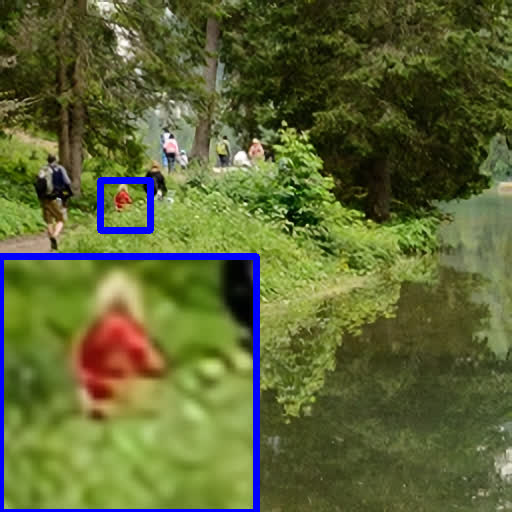}&
\includegraphics[width=0.2\linewidth]{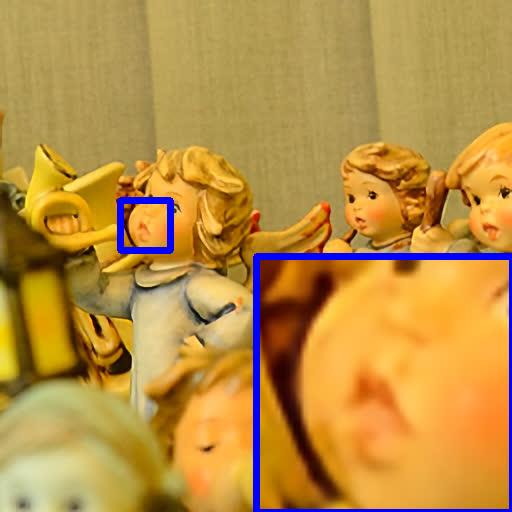}&
\includegraphics[width=0.2\linewidth]{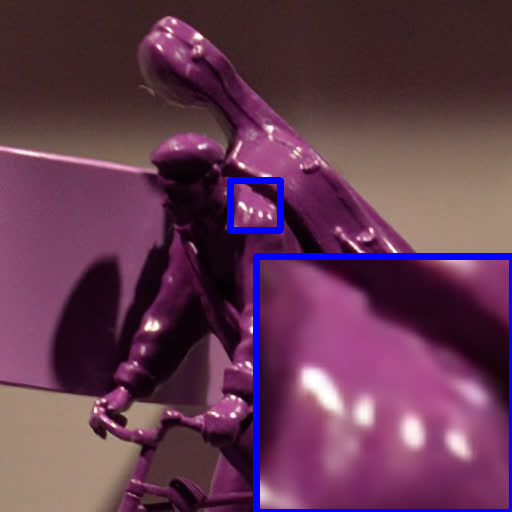}\\
(25.8677dB , 0.9259)&(28.1375 dB , 0.9468)&(34.0613dB , 0.9950)&(38.0707dB , 0.9892)
\end{tabular}
\caption{Joint demosacing and super-resolution results on images from the \textbf{RAISE}~\cite{DangNguyen:2015va} dataset. The two numbers in the brackets are the PSNR and SSIM scores, respectively.}
\label{fig:result2}
\end{figure*}
Since we are not aware of any other joint demosaicing and super-resolution algorithms in existing literature, we compare our method with the sequential application of different state-of-the-art demosaicing algorithms(FlexISP~\cite{Pulli:2014gq}, SEM~\cite{Klatzer:2016wg} and DemosaicNet~\cite{Durand:2016tz}) and the state-of-the-art super-resolution algorithm(SRCNN~\cite{Dong:2014vh}).

Note that SEM~\cite{Klatzer:2016wg} and DemosaicNet~\cite{Durand:2016tz} perform joint demosaicing and denoising, for fair comparison, we set noise-level $= 0$ for these methods. As SRCNN only provides upsampling in the luminance channel, we upsample the chroma channels using bicubic interpolation. The process is shown in Figure. ~\ref{fig:diagram}. We use the 9-5-5 model of SRCNN.

\subsubsection{Quantitative Results}
In Table.~\ref{tab:result} we report the PSNR values of our approach in comparison to other methods on the testing dataset. Our approach outperforms the PSNR scores  of the next best combination of state-of-the-art techniques of demosaicing and super-resolution by a significant PSNR difference of $1.3 dB$ on the average computed over the 50 images of the test-set.
\begin{table}[h]
    \begin{tabular}{|c|c|c|}
        \hline
         Method & PSNR & SSIM  \\
        \hline
        FlexISP~\cite{Pulli:2014gq}+SRCNN~\cite{Dong:2014vh} & 29.6092 dB & 0.9182\\
        SEM*~\cite{Klatzer:2016wg}+SRCNN~\cite{Dong:2014vh} & 29.4978 dB & 0.9348\\
        DemosaicNet*~\cite{Durand:2016tz}+SRCNN~\cite{Dong:2014vh} & 30.1313 dB & 0.9374\\
        Ours & \textbf{31.4093 dB} & \textbf{0.9476}\\
        \hline
    \end{tabular}
    \caption{The mean PSNR and SSIM of different methods evaluated on our testing dataset. For the methods that perform joint demosaicing and denoising, we set their noise-level to 0 for fair comparison. There is a significant difference between the PSNRs and SSIMs of our proposed network and existing state-of-the-art methods.}
    \label{tab:result}
\end{table}

\subsubsection{Qualitative Results}
To further validate the quality of our results, we show qualitative comparisons in Figure.~\ref{fig:result1} and Figure.~\ref{fig:result2}. 

The combination of FlexISP~\cite{Pulli:2014gq} and SEM~\cite{Klatzer:2016wg} produces some disturbing artifacts such as zippering around the edge and false color artifacts. These are particularly visible in the man's clothes (in the first column of Figure.~\ref{fig:result1}) and the text (in the last column of Figure.~\ref{fig:result2}).

Both DemosaicNet~\cite{Durand:2016tz} and our network can produce demosaiced images without these artifacts, but our network is able to recover more realistic details. This is demonstrated in the first and the third column of Figure.~\ref{fig:result2}. Our network is able to produce higher quality color images without the visually disturbing artifacts introduced by the other methods.

\subsubsection{Running Time}
We test the running time of DemosaicNet~\cite{Durand:2016tz} and our method on 10 $256 \times 256$ input images using a Nvidia TITAN X. While DemosaicNet takes on average 650 ms for demosaicing alone, our method has an average of 619 ms for the joint operation of demosaicing and super-resolution.

%% file: texinput/conclusion.tex
\section{Conclusion}
The ill-posed problems of demosaicing and super-resolution have always been dealt with as separate problems and then applied sequentially to obtain high-resolution images. This has continued to remain the trend even after the advent of CNN's. In this paper, for the first time as far as we know, we propose a CNN-based joint demosaicing and super-resolution framework, which is capable of directly recovering high-quality color super-resolution images from Bayer mosaics. Our approach does not produce disturbing color artifacts akin to algorithms in the literature. Our proposed method outperforms all the tested combinations of the state-of-the-art demosaicing algorithms and the state-of-the-art super-resolution algorithms in both quantitative measurements of PSNR and SSIM as well as visually. This augurs well for the use of our approach in camera image processing pipelines. For mobile devices this can encourage the use of sensors with large pixels that capture a better dynamic range, rather than sacrificing dynamic range for higher resolution as is done at the moment.

%% file: main.bbl
\begin{thebibliography}{10}\itemsep=-1pt

\bibitem{Allebach:1996uz}
J.~P. Allebach and P.~W. Wong.
\newblock {Edge-directed interpolation.}
\newblock {\em ICIP}, 1996.

\bibitem{Alley:2005ez}
D.~Alleysson, S.~Susstrunk, and J.~Herault.
\newblock {Linear demosaicing inspired by the human visual system}.
\newblock {\em IEEE Transactions on Image Processing}, 14(4):439--449, 2005.

\bibitem{DangNguyen:2015va}
D.-T. Dang-Nguyen, C.~Pasquini, V.~Conotter, and G.~Boato.
\newblock {RAISE - a raw images dataset for digital image forensics.}
\newblock {\em MMSys}, 2015.

\bibitem{xtrans}
J.~H. David~Alleysson, Brice Chaix De~Lavarene.
\newblock {Digital image sensor, image capture and reconstruction method and
  system for implementing same.}
\newblock {\em US patent}, 2012.

\bibitem{Dong:2014vh}
C.~Dong, C.~C. Loy, K.~He, and X.~Tang.
\newblock {Learning a Deep Convolutional Network for Image Super-Resolution.}
\newblock {\em ECCV}, 2014.

\bibitem{Dong:2016ek}
C.~Dong, C.~C. Loy, and X.~Tang.
\newblock {Accelerating the Super-Resolution Convolutional Neural Network}.
\newblock {\em ECCV}, 2016.

\bibitem{Egiazarian:2015ww}
K.~O. Egiazarian and V.~Katkovnik.
\newblock {Single image super-resolution via BM3D sparse coding.}
\newblock {\em EUSIPCO}, 2015.

\bibitem{Elad:2010wu}
M.~Elad.
\newblock {Sparse and Redundant Representations - From Theory to Applications
  in Signal and Image Processing.}
\newblock 2010.

\bibitem{Farsiu:2006ug}
S.~Farsiu, M.~Elad, and P.~Milanfar.
\newblock {Multiframe demosaicing and super-resolution of color images.}
\newblock {\em IEEE Trans. Image Processing}, 2006.

\bibitem{Freedman:2011cf}
G.~Freedman and R.~Fattal.
\newblock {Image and video upscaling from local self-examples.}
\newblock {\em ACM Transactions on Graphics}, 30(2):1--11, 2011.

\bibitem{Smith:2012hv}
D.-Y.~Y. H.~Chang and Y.~Xiong.
\newblock {Super-Resolution of Text Images through Neighbor Embedding.}
\newblock {\em CVPR}, 2012.

\bibitem{Tan:2017}
S.~L. Hanlin~Tan, Xiangrong~Zeng and M.~Zhang.
\newblock {Joint Demosaicing and Denoising of Noisy Bayer Images with ADMM}.
\newblock {\em IEEE International Conference on Image Processing}, 2016.

\bibitem{He:2016tt}
K.~He, X.~Zhang, S.~Ren, and J.~Sun.
\newblock {Deep Residual Learning for Image Recognition.}
\newblock {\em CVPR}, 2016.

\bibitem{Hirakawa:cy}
K.~Hirakawa and T.~W. Parks.
\newblock {Adaptive homogeneity-directed demosaicing algorithm}.
\newblock {\em IEEE Transactions on Image Processing}, (3):360--469, 2005.

\bibitem{Kim:2016kv}
J.~K.~L. J.~Kim and K.~M. Lee.
\newblock {Deeply-Recursive Convolutional Network for Image Super-Resolution.}
\newblock {\em CVPR}, 2016.

\bibitem{Sun:2008ia}
Z.~X. J.~Sun and H.-Y. Shum.
\newblock {Image super-resolution using gradient profile prior.}
\newblock {\em CVPR}, 2008.

\bibitem{Glotzbach:jm}
K.~I. J.W.~Glotzbach, R.W.~Schafer.
\newblock {A method of color filter array interpolation with alias cancellation
  properties}.
\newblock {\em International Conference on Image Processing}, 2002.

\bibitem{He:2015cv}
S.~R. Kaiming~He, Xiangyu~Zhang and J.~Sun.
\newblock {Delving Deep into Rectifiers: Surpassing Human-Level Performance on
  ImageNet Classification.}
\newblock {\em IEEE International Conference on Computer Vision}, 2015.

\bibitem{Kim:2016wv}
J.~Kim, J.~K. Lee, and K.~M. Lee.
\newblock {Accurate Image Super-Resolution Using Very Deep Convolutional
  Networks.}
\newblock {\em CVPR}, 2016.

\bibitem{Kingma:2014us}
D.~P. Kingma and J.~Ba.
\newblock {Adam: A Method for Stochastic Optimization.}
\newblock {\em International Conference for Learning Representations}, 2014.

\bibitem{Klatzer:2016wg}
T.~Klatzer, K.~Hammernik, P.~Kn{\"o}belreiter, and T.~Pock.
\newblock {Learning joint demosaicing and denoising based on sequential energy
  minimization.}
\newblock {\em ICCP}, 2016.

\bibitem{Krizhevsky:2012wl}
A.~Krizhevsky, I.~Sutskever, and G.~E. Hinton.
\newblock {ImageNet Classification with Deep Convolutional Neural Networks.}
\newblock {\em NIPS}, 2012.

\bibitem{:2001uc}
X.~Li and M.~T. Orchard.
\newblock {New edge-directed interpolation.}
\newblock {\em IEEE Trans. Image Processing}, 2001.

\bibitem{Lim:2017un}
B.~Lim, S.~Son, H.~Kim, S.~Nah, and K.~M. Lee.
\newblock {Enhanced Deep Residual Networks for Single Image Super-Resolution.}
\newblock {\em CVPR Workshops}, 2017.

\bibitem{Mao:2016ti}
X.-J. Mao, C.~Shen, and Y.-B. Yang.
\newblock {Image Restoration Using Very Deep Convolutional Encoder-Decoder
  Networks with Symmetric Skip Connections.}
\newblock {\em NIPS}, 2016.

\bibitem{Durand:2016tz}
D.~S.~P. Michael~Gharbi, Gaurav~Chaurasia and Fr{\'e}do.
\newblock {Deep joint demosaicking and denoising.}
\newblock {\em ACM Trans. Graph.}, 2016.

\bibitem{Pulli:2014gq}
K.~Pulli.
\newblock {FlexISP - a flexible camera image processing framework.}
\newblock {\em ACM Trans. Graph.}, 33(6):1--13, 2014.

\bibitem{Schulter:2015tz}
S.~Schulter, C.~Leistner, and H.~Bischof.
\newblock {Fast and accurate image upscaling with super-resolution forests.}
\newblock {\em CVPR}, 2015.

\bibitem{Shi:2016tm}
{Shi, Wenzhe}, {Caballero, Jose}, {Huszar, Ferenc}, {Totz, Johannes}, {Aitken,
  Andrew P}, {Bishop, Rob}, {Rueckert, Daniel}, and {Wang, Zehan}.
\newblock {Real-Time Single Image and Video Super-Resolution Using an Efficient
  Sub-Pixel Convolutional Neural Network.}
\newblock {\em CVPR}, 2016.

\bibitem{Timofte:2015kw}
R.~Timofte, V.~De~Smet, and L.~Van~Gool.
\newblock {A : Adjusted Anchored Neighborhood Regression for Fast
  Super-Resolution.}
\newblock {\em ACCV}, 2014.

\bibitem{Wu:hw}
X.~Wu and N.~Zhang.
\newblock {Primary-consistent soft-decision color demosaic for digital
  cameras}.
\newblock {\em IEEE Transactions on Image Processing}, (9):1263--1274, 2004.

\bibitem{mao:2016}
C.~S. X.~Mao and Y.-B. Yang.
\newblock {Image Restoration Using Very Deep Convolutional Encoder-Decoder
  Networks with Symmetric Skip Connections.}
\newblock {\em NIPS}, 2016.

\bibitem{Li:2008ic}
B.~G. Xin~Lia and L.~Zhang.
\newblock {Image demosaicing: a systematic survey.}
\newblock {\em Visual Communications and Image Processing}, 2008.

\bibitem{Yang:2012db}
J.~Yang, Z.~Wang, Z.~Lin, S.~Cohen, and T.~Huang.
\newblock {Coupled Dictionary Training for Image Super-Resolution}.
\newblock {\em IEEE Transactions on Image Processing}, 21(8):3467--3478, 2012.

\end{thebibliography}
